\def\year{2019}\relax
\documentclass[letterpaper]{article} %DO NOT CHANGE THIS
\usepackage{aaai19}  %Required
\usepackage{times}  %Required
\usepackage{helvet}  %Required
\usepackage{courier}  %Required
\usepackage{url}  %Required
\usepackage{graphicx}  %Required
\begin{document}
\title{A Comparison Study of Credit Card Fraud Detection: \\Supervised versus Unsupervised}
\author{Xuetong Niu, Li Wang, Xulei Yang*\\
%Association for the Advancement of Artificial Intelligence\\
%2275 East Bayshore Road, Suite 160\\
%Palo Alto, California 94303\\
}
\maketitle
\begin{abstract}
Credit card has become popular mode of payment for both online and offline purchase, which leads to increasing daily fraud transactions. An Efficient fraud detection methodology is therefore essential to maintain the reliability of the payment system. In this study, we perform a comparison study of credit card fraud detection by using various supervised and unsupervised approaches. Specifically, 6 supervised classification models, i.e., Logistic Regression (LR), K-Nearest Neighbors (KNN), Support Vector Machines (SVM),  Decision Tree (DT), Random Forest (RF), Extreme Gradient Boosting (XGB), as well as 4 unsupervised anomaly detection models, i.e., One-Class SVM (OCSVM), Auto-Encoder (AE), Restricted Boltzmann Machine (RBM), and Generative Adversarial Networks (GAN), are explored in this study. We train all these models on a public credit card transaction dataset from Kaggle website, which contains 492 frauds out of 284,807 transactions. The labels of the transactions are used for supervised learning models only. The performance of each model is evaluated through 5-fold cross validation in terms of Area Under the Receiver Operating Curves (AUROC). Within supervised approaches, XGB and RF obtain the best performance with AUROC = 0.989 and AUROC = 0.988, respectively. While for unsupervised approaches, RBM achieves the best performance with AUROC = 0.961, followed by GAN with AUROC = 0.954. The experimental results show that supervised models perform slightly better than unsupervised models in this study. Anyway, unsupervised approaches are still promising for credit card fraud transaction detection due to the insufficient annotation and the data imbalance issue in real-world applications. 
\end{abstract}
\section{Introduction}
Credit card fraud detection has recently become an active research topic with the exploting growth of big data and AI techniques. Also, it plays an important role in banks as it would help to reduce loss caused by fraudulent transactions. Although many proposed methods \cite{zareapoor2015application,randhawa2018credit} have achieved promising results, it is still very challenging to accurately and promptly detect credit card fraudulent transactions due to dramatic data imbalance and large variations of fraud transactions.

Both supervised and unsupervised learning have been investigated in credit card fraud detection. For example, a combination of multiple learned fraud detectors \cite{chan1999distributed} is proposed under a so-called ``cost model" to solve the problem of skewed distribution for training data. In contrast, an unsupervised method \cite{bolton2001unsupervised} is proposed to detect changes in behavior of usual credit card transactions rather than relying on labels of fraudulent historical transaction data. Also, some surveys have comprehensively studied machine learning techniques applied to credit card fraud detection. For example, the survey \cite{zojaji2016survey} reviews the techniques, datasets and evaluation criteria in credit card fraud detection. However, no one has evaluated machine learning models and compared credit card fraud detection performance in a supervised vs unsupervised manner.

In this paper, we evaluate $5$ supervised learning models and 4 unsupervised learning models on a Kaggle credit card transaction dataset. The supervised learning models include Support Vector Machines (SVM) \cite{cortes1995support}, K-Nearest Neighbors (KNN) \cite{altman1992introduction}, Extreme Gradient Boosting (XGB) \cite{chen2015xgboost}, Logistic Regression (LR) \cite{neter1996applied}, Decision Tree (DT) \cite{quinlan1986induction} and Random Forest (RF) \cite{breiman2001random}, while the unsupervised learning methods contain One-Class SVM (OCSVM) \cite{scholkopf2000support}, Auto-Encoder (AE) \cite{deng2010binary}, Restricted Boltzmann Machine (RBM) \cite{sutskever2009recurrent}, and Generative Adversarial Networks (GAN) \cite{goodfellow2014generative}. The supervised learning models leverage transaction labels to train classifiers that are able to distinguish between normal and abnormal transactions. In contrast, the unsupervised learning models use unlabeled data for training to capture normal data distribution and then determine whether an unknown test sample is normal or abnormal. As labeling data is time-consuming and labor intensive, labeled data is very expensive, especially when abnormal samples are much smaller than normal one. In this case, the unsupervised learning models would be more useful than the supervised one.

The main contribution of this paper is that we comprehensively studied both supervised and unsupervised learning models for credit card fraud detection and evaluate these machine learning algorithms on a Kaggle credit card transaction dataset in a supervised vs unsupervised way. According to our best knowledge, we are the first to conduct this sort of comparison study between supervised and unsupervised learning on credit card fraud detection.

\section{Related Works}
\subsection{Traditional Machine Learning Methods}
It is very time-consuming for people to check credit card transactions one-by-one as transaction amount is tremendously large. Hence, an automated method is desired for credit card fraud detection. In decades, many machine learning methods have been used to solve this problem. Next, we will review some of them to have a big picture of this research area. The traditional neural networks (compared to the current deep neural networks) have already been used for credit card fraud detection in \cite{dorronsoro1997neural}. Hidden Markov Model (HMM) \cite{srivastava2008credit} is utilized to model the sequence of operations in credit card transaction processing and detect frauds. In \cite{bhattacharyya2011data}, Support Vector Machine (SVM) and Random Forest (RF) are investigated together with Logistic Regression (LR) based on real-life data from international credit card transactions. Also, a cost-sensitive decision tree based method \cite{sahin2013cost} is proposed for credit card fraud detection and evaluated on a real world dataset. In another work \cite{mahmoudi2015detecting}, a modified Fisher discriminant function is proposed for credit card fraud detection to be more sensitive to important instances. Besides using machine learning methods, a framework for transaction aggregation \cite{whitrow2009transaction} is proposed to solve the problem of preprocessing credit card transaction data for supervised fraud classification. Also, a novel learning strategy \cite{dal2018credit} is proposed to solve three issues of class imbalance, concept drift and verification latency in credit card fraud detection.

\subsection{Advanced Deep Learning Methods}
Recently, deep learning algorithms have achieved promising results in many areas such as image processing \cite{wang2015video}. Therefore, we will review several deep learning based works for credit card fraud detection as follows. Long Short-Term Memory (LSTM) is utilized in \cite{DBLP:journals/eswa/JurgovskyGZCPHC18} to formulate the credit card fraud detection as a sequence classification problem belonging to supervised learning. Also, an unsupervised model \cite{Pumsirirat2018} of deep Auto-Encoder (AE) and Restricted Boltzmann Machine (RBM) is proposed to reconstruct credit card normal transactions and detect anomalies. Specifically, a framework tuning parameters of deep learning topologies is proposed for credit card fraud detection in \cite{roy2018deep}. It is necessary to mention that Generative Adversarial Network (GAN) is a remarkable model in unsupervised and semi-supervised learning. Not only it is employed to detect activity fraud and malicious users in online social networks \cite{DBLP:journals/corr/abs-1803-01798}, but also it has been used in credit card fraud detection \cite{fiore2017using} to augment minority class examples for the classification between fraudulent and non-fraudulent samples. In this paper, the GAN model will also be studied and evaluated as one of unsupervised learning methods.

\section{Supervised Learning Methods}

Some machine learning methods treat fraud transaction as a supervised classification problem. In this way, we can train a classifier based on training data together with annotations, then classify test transaction data into normal and abnormal categories. In this Section, we briefly discuss 6 widely-used supervised machine learning approaches for credit card fraud detection.

\subsection{Logistic Regression}
Logistic regression was developed by statistician David Cox in 1958 and is a regression model where the response variable Y is categorical. Logistic regression allows us to estimate the probability of a categorical response based on one or more predictor variables $x$. It allows one to say that the presence of a predictor increases (or decreases) the probability of a given outcome by a specific percentage.  Mathematically, logistic regression estimates a multiple linear regression function defined as:
\begin{eqnarray}\label{LR}
Y_i = \beta_0 + \beta_1x_{i,1} + \beta_2x_{i,2} + ... + \beta_px_{i,p}
\end{eqnarray}
where $x_{i,j}$ refers to the $j^{th}$ predictor variable for the $i^{th}$ observation, $Y_i$ is the output of $i^{th}$ observation.

\subsection{K-Nearest Neighbors}
In the classification setting, the KNN algorithm essentially boils down to forming a majority vote between the $K$ most similar instances to a given “unseen” observation. Similarity is defined according to a distance metric between two data points $x$ and $x'$ . A popular choice is the Euclidean distance given by
\begin{eqnarray}
d(x,x')=\sqrt{(x_1 - {x_1}')^2 + (x_2 - {x_2}')^2+ ... + (x_n - {x_n}')^2}
\end{eqnarray}
But other measures can be more suitable for a given setting and include the Manhattan, Chebyshev and Hamming distance.
More formally, given a positive integer K, an unseen observation $x$ and a similarity metric $d$, KNN classifier performs the following two steps: It runs through the whole dataset computing $d$ between $x$ and each training observation. Suppose the $K$ points in the training data that are closest to $x$ are denoted as set $A$. 
It then estimates the conditional probability for each class, that is, the fraction of points in $A$ with that given class label. 
\begin{eqnarray}
P(y=j|X=x) = \frac{1}{K}\sum_{i \in A} I(y^i = j)
\end{eqnarray}
where $I(x)$ is the indicator function which evaluates to $1$ when the argument $x$ is true and 
$0$ otherwise Finally, the input $x$ is assigned to the class with the largest probability.

\subsection{Support Vector Machine}
SVM was first introduced by Vapnik in 1995 to solve the classification and regression problems. 
The basic idea of SVM is to derive an optimal hyperplane that maximizes the margin between two classes. 
A nice property of SVMs is that it can find a non-linear decision boundary by projecting the data through a nonlinear function $\phi$ to a space with a higher dimension. This means that data points which cannot be separated by a straight line in their original input space are lifted to a feature space $F$ where there can be a linear hyperplane separating the data points of one class from an other. When that hyperplane would be projected back to the input space I, it would have the form of a non-linear curve.

Mathematically, given $n$ training data samples
\[
\{(x_i, y_i)\}_{i=1}^n, \hspace{.1in} x_i\in R^N, y_i\in\{-1,1\}
\]
SVM is formulated by the following optimization problem:
\begin{eqnarray}\label{svm_l}
Minimize \hspace{.1in} \Phi(w)=\frac{1}{2}w^Tw+C\sum_{i=1}^{n}
\xi_i
\end{eqnarray}
subject to
\begin{eqnarray}\label{svm_c}
y_i(\langle w,\phi(x_i)\rangle+b)\geq{1-\xi_i}, \hspace{.1in} i=1,\dots,n \nonumber \\
\xi_i \geq 0, \hspace{.1in} i=1,\dots,n
\end{eqnarray}
where the kernel function $\phi$ maps training points $x_i$ from input space into a higher
dimensional feature space. The regularization parameter $C$ controls the trade-off between achieving a low error on the training data and minimising the norm of the weights. . 

\subsection{Decision Tree}
Decision trees are simple but intuitive models that utilize a top-down approach in which the root node creates binary splits until a certain criteria is met. This binary splitting of nodes provides a predicted value based on the interior nodes leading to the terminal (final) nodes. In a classification context, a decision tree will output a predicted target class for each terminal node produced. 

Decision trees tend to have high variance when they utilize different training and test sets of the same data, since they tend to overfit on training data. This leads to poor performance on unseen data. Unfortunately, this limits the usage of decision trees in predictive modeling. However, using ensemble methods, we can create models that utilize underlying decision trees as a foundation for producing powerful results.

\subsection{Random Forest}

The random forest algorithm, proposed by L. Breiman in 2001, has been successful as a general-purpose classification and regression method. The approach, which combines several randomized decision trees and aggregates their predictions by averaging, has shown excellent performance in the setting where the number of variables is much larger than the number of observations. Moreover, it is versatile enough to be applied to large-scale problems, is easily adapted to various ad-hoc learning tasks, and returns measures of variable importance.

In the classification context, the random forest classifier $m$ is obtained via a majority vote among $K$ classification trees with input $x$, that is, 
\begin{eqnarray}
m(x:\Theta_1, ... , \Theta_K)= 
\lbrace
\begin{array}{rcl}
1   &      &  if \frac{1}{K}\sum_{j=1}^K m(x; \Theta_j) > \frac{1}{2} \\
0   &      &  otherwise
\end{array} 
\end{eqnarray}
where $\Theta$ is the parameter set. 

%$$ m_K(x:\Theta_1, ... , \Theta_K)=\left\{
%\begin{array}{rcl}
%1   &      &  if \frac{1}{K}\sum_{j=1}^K m(x; \Theta_j) > \frac{1}{2} \\
%0   &      &  otherwise
%\end{array} \right. $$

\subsection{Extreme Gradient Boosting}
Gradient boosting is a powerful machine learning technique for regression, classification and ranking problems, which produces a prediction model in the form of an ensemble of weak prediction models like decision trees. The model is built in a stage-wise manner. In each stage, it introduces a new weak learner to compensate the
shortcomings of the existing weak learners. XGB stands for eXtreme Gradient Boosting, one of the implementations of gradient boosting concept. The unique of XGB is that it uses a more regularized model formalization to control over-fitting and achieves better performance.

Gradient boosting relies on regression trees, where the optimization step works to reduce mean square error, and for binary classification the standard log loss is used. For a multi-class classification problem, the objective function is to optimize the cross entropy loss. Combining the loss function with a regularization term arrives at the objective function. The regularization term controls the complexity and reduces the risk of over-fitting. XGB uses gradient descent for optimization to improve the predictive accuracy at each optimization step by following the negative of the gradient as we are trying to find the “sink” in a n-dimensional plane. 

To learn the set of functions used in the model, XGB minimizes the following regularized objective
\begin{eqnarray}
L(\Theta) = \sum_{i} l(y_i, \hat y_i) +  \Omega (\Theta)
\end{eqnarray}
where $\Theta$ is the learned parameter set, $l$ is a differentiable convex loss function that measures the difference between the predictions $\hat y_i$ and the target $y_i$, and $\Omega$ is the regularization term.

\section{Unsupervised Learning Methods}

There is a recent surge of interest in developing unsupervised generative models for anomaly detection. Generative models are trained to model the distribution of the normal transaction data (without annotations) distribution. Any transaction that does not follow the distribution is considered to be anomalous. In such a way, the fraud transaction can be detected in an unsupervised manner. In this Section, we briefly discuss 4 unsupervised machine learning approaches for credit card fraud detection. 

\subsection{One-Class Support Vector Machine}

One-Class SVM (OCSVM) was proposed by scholkopf to identify novelty / anomaly in an unsupervised manner without labeled training data. The algorithm learns a soft boundary in order to embrace the normal data instances using the training set, and then, using the testing instance, it tunes itself to identify the abnormalities that fall outside the learned region.

Mathematically, OCSVM is formulated by the following optimization problem :
\begin{eqnarray}\label{svm_l}
Minimize \hspace{.1in} \Phi(w)=\frac{1}{2}w^Tw+\frac{1}{\upsilon n}\sum_{i=1}^{n}\xi_i - \rho 
\end{eqnarray}
subject to
\begin{eqnarray}\label{svm_c}
y_i(\langle w,\phi(x_i)\rangle+b)\geq{\rho -\xi_i}, \hspace{.1in} i=1,\dots,n \nonumber \\
\xi_i \geq 0, \hspace{.1in} i=1,\dots,n
\end{eqnarray}
The parameter $\upsilon$ sets an upper bound on the fraction of outliers and a lower bound on the number of training examples used as support vectors.

\subsection{Restricted Boltzmann Machine}
A RBM model consists of visible and hidden layers, which are connected through symmetric weights. The
inputs $x$ correspond to the neurons in the visible layer. The response of the neurons $h$ in the hidden layer model the probability distribution of
the inputs. The probability distribution is derived by learning
the symmetrical connecting weights between the visible and the hidden layers. The neurons in the same layer are not connected.
The conditional probability of a configuration of the hidden
neurons (h), given a configuration of the visible neurons
associated with inputs (x), is:
\begin{eqnarray}
p(h|x) = \prod_i{p(h_i|x)}
\end{eqnarray}
The objective of the generative training in RBM is to learn
the unknown (h) iteratively using the input (x).

The generative training phase iterates until the reconstructed samples most
closely approximates $x$. It is performed using the
maximum likelihood criterion, and implemented by minimizing the
negative log probability of the training data:
\begin{eqnarray}
L_{gen} = -\sum logP(x|(w_{ij}, b_i, c_j))
\end{eqnarray}
where $b_i$ and $c_j$ are the bias in the input and hidden layers,
respectively. $w_{ij}$ denotes the weights between the inputs and hidden layers.

\subsection{Auto-Encoder}
An auto-encoder (AE) learns to map from input to output through a pair of encoding and decoding phases. The encoder maps from the input to hidden layer, the decoder maps from the hidden layers to the output layer to reconstruct the inputs. The hidden layers of the auto-encoder are low-dimensional and nonlinear representation of the input
data. 
The AE is formulated as follows,
\begin{eqnarray}
\hat{X} = D(E(X))
\end{eqnarray}
where $X$ is the input data, $E$ is an encoding map, $D$ is a
decoding map, and $\hat{X}$ is the reconstructed input data.
The objective of the auto-encoder is to approximate the
distribution of $X$ as accurately as possible. In particular, an
autoencoder can be viewed as a solution to the following
optimization problems:
\begin{eqnarray}
min_{D,E} \|X-D(E(X))\|
\end{eqnarray}
where $\|\cdot\|$ is usually 2-norm. Complex distributions of $X$ can
be modelled using a deep auto-encoder with multiple layers,
which refers to multiple pairs of encoders and decoders.

\subsection{Generative Adversarial Networks}
GAN is a generative model designed by Goodfellow in 2014. In a GAN setup, two differentiable functions (generator $G$ and discriminator $D$), represented by neural networks, are competing and trained simultaneously, which eventually drive the generated samples to be indistinguishable from real data. 

The GAN model in this study is based on AnoGAN \cite{schlegl2017unsupervised} recently developed for anomaly detection by T. Schlegl etc.  We modified the original AnoGAN by simultaneously learn an encoder $E$ that maps input samples x to a latent representation z, along with a generator $G$ and discriminator $D$ during training. This enables us to avoid the computationally expensive SGD step for recovering a latent representation at test time.  

After we train the model on the normal data to yield $G$, $D$ and $E$ for inference, we also define a score function $A(x)$ that measures how anomalous an example $x$ is, based on a convex combination of a reconstruction loss $L_G$ and a discriminator-based loss $L_D$:
\begin{eqnarray} 
A(x) = \alpha*L_G(x) + (1-\alpha)*L_D(x)
\end{eqnarray}

where 
\[
L_G(x) = ||x - G(E(x))||_1 
\]
and
\[ 
L_D(x) = \sigma (D(x, E(x)),1)
\] 
where $\alpha$ is a weighting parameter ranged in $(0,1)$, $\sigma$ is the cross-entropy loss from the discriminator of $x$ being a real example (class $1$). The definition of $L_G(x)$ indicates how well the trained encoder and generator can reconstruct an input example $x$. The definition of $L_D(x)$ captures the discriminator confidence that a sample is derived from the real data distribution. 

\section{Experimental Results}

\subsection{Data Set and Preprocessing}
\begin{figure}[!htbp]
	\centering
	\includegraphics[width=8cm]{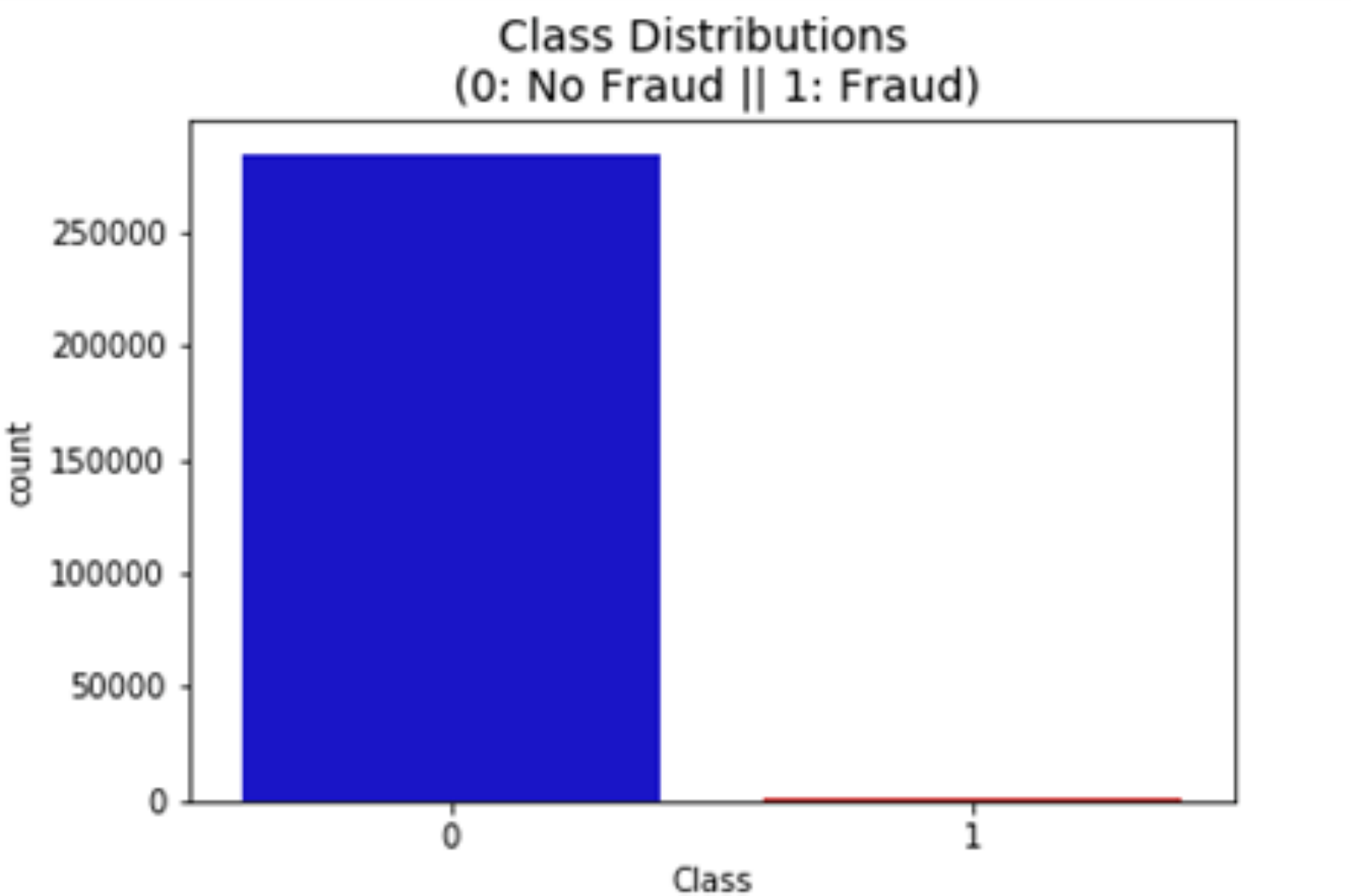}
	\caption{Number of Different Classes}\label{lbl:dataset}
\end{figure}

This public dataset contains credit card transactions made in September 2013 by European cardholders. The transactions occurred in two days include 492 fraud records out of 284,807 transactions. It is obvious that the dataset is highly unbalanced (Fig.\ref{lbl:dataset}). The fraudulent class only accounts for $0.172\%$ of all transactions. 

The dataset contains numerical input variables which are from a PCA transformation due to confidentiality issue. For the non-numerical features of ``Time" and ``Amount", we normalize them by using RobustScaler which scales the data according to the quantile range. Specifically for the supervised learning models, to tackle the heavily unbalanced problem, random downsampling is used to avoid the bias results toward the non-fraudulent class. Through random downsampling, non-fraud transactions (Class = ‘0’) are randomly reduced to the same amount as fraud transactions (Class = ‘1’), which is equivalent to 492 cases of frauds and 492 cases of non-fraud transactions. 

\subsection{Evaluation Metrics} 
As mentioned above, the studied data set is highly imbalanced with 492 fraud records out of 284,807 transactions. Even all the samples are classified into non-fraud category, the classification accuracy is still extremely high, that means traditional evaluation metrics like accuracy is not suitable for this study. Instead, we report the Area Under the Receiver Operating Curves (AUROC) \cite{} in our experimental study. AUROC combines the false positive rate (FPR) and the true positive rate (TPR) into one single metric. With the assumption that fraud class is ``positive" and non-fraud class is ``negative”, the definition of FPR and TPR are as follows:
 \[TPR = TP / P\]
and 
 \[FPR = FP / N\]
where P and N are the number of samples from “positive” and “negative” classes, respectively.  TP (True Positive) represents the number of samples predicted to be “positive” while they are actually positive, and FP (False Positive) the number of samples predicted to be “positive” while they are actually negative.

To avoid overfitting issues, in this study, $k$-fold cross-validation technique is used to estimate fraud detection performance. In one round of k-fold cross-validation, the data set is first randomly divided into $k$ subsets (or folds), which are of approximately equal size and are mutually exclusive. A machine learning model is then trained and tested $k$ times, where in each time, one of the subsets is set aside as the testing data and the remaining $k − 1$ subsets are used as training data. The final testing results are predicted from $k$ trained sub-models. In our experimental studies, 5 cross validations (i.e., $k = 5$) are used as the validation method.

\subsection{Parameter Settings}
The key parameters of most studied models are determined by grid-search through cross validation, which are listed below:
\begin{itemize}
	\item LR:  {'C': 0.1, 'penalty': 'l1'}
	\item KNN:  {'algorithm': 'auto', 'n\_neighbors': 4}
	\item SVM:  {'C': 0.5, 'kernel': 'linear'}
	\item DT:  {'criterion': 'entropy', 'max\_depth': 3, 'min\_samples\_leaf': 6}
	\item RF:  {'n\_estimators': 30, 'oob\_score': 'True'}
	\item XGB:   {'learning\_rate': 0.4, 'max\_depth': 4}
	\item OCSVM: {'nu': 0.1, 'gamma': 0.001}
	\item RBM: {'learning\_rate':  0.0005 'num\_hidden':  10} 
\end{itemize}
While the neural network architectures for Auto-encoder and Generative Adversarial Networks are shown below:
\begin{itemize}
	\item AE: The encoder has two dense layers with 16 and 32 Relu units, each. 
	The decoder has two dense layers of 32 and 16 Relu units, respectively. 
	\item GAN: The encoder has two dense layers with 32 leaky ReLu and 32 linear units, each. The
	generator has three dense layers of 32 ReLu, 64 ReLu and
	28 linear units, respectively. And the discriminator has one dense
	layer of 32 leaky ReLu units followed by one linear
	layer with single unit.
\end{itemize}

\subsection{Results}
\begin{figure*}[!htbp]
	\centering
	\includegraphics[width=0.9\linewidth]{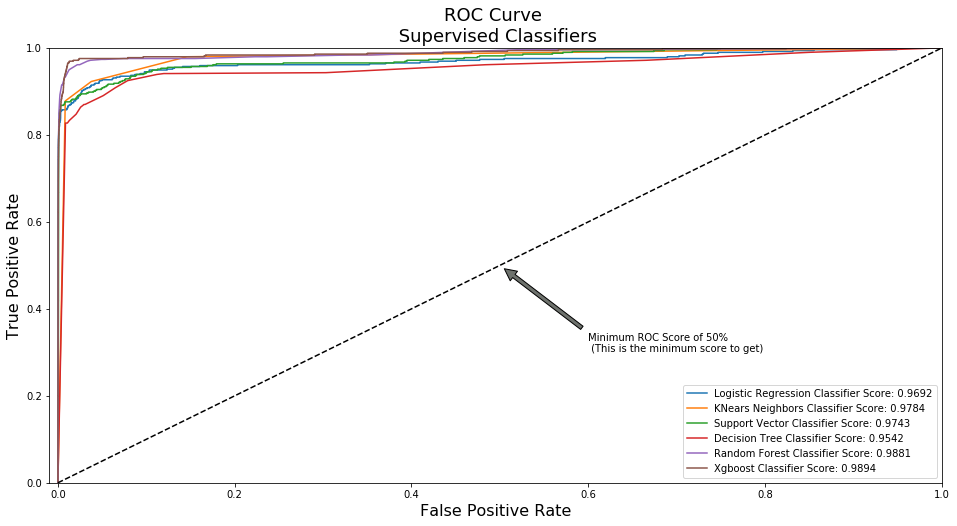}
	\caption{Plot of AUROC by supervised approaches}\label{lbl:supervised}
\end{figure*}

\begin{figure*}[!htbp]
	\centering
	\includegraphics[width=0.9\linewidth]{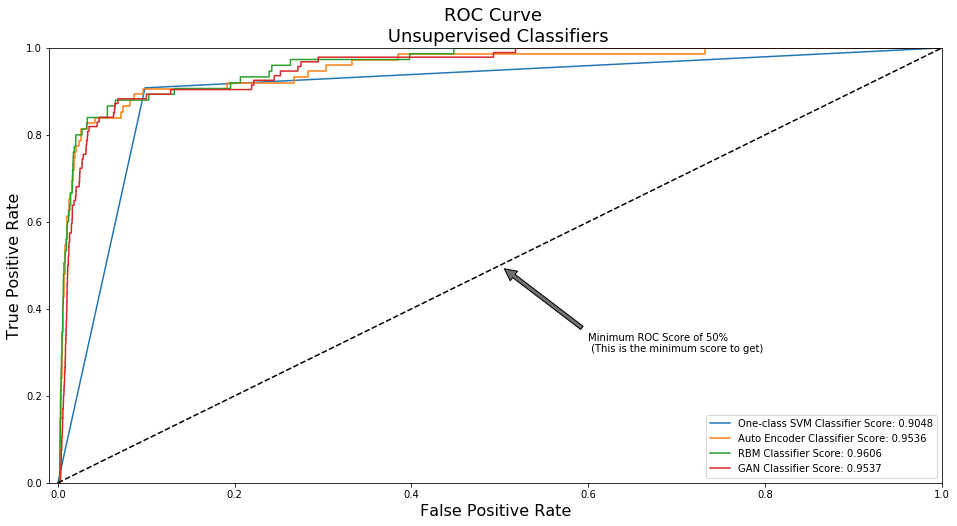}
	\caption{Plot of AUROC by unsupervised approaches}\label{lbl:unsupervised}
\end{figure*}

The AUROC values of the 6 supervised models on the studied credit card transaction dataset are shown in Fig.\ref{lbl:supervised}. It can be seen that all the models perform well on this data set, with XGB achieves the best performance with AUROC=0.99, while DT obtains the lowest AUROC value of 0.95. It is expected that the ensemble methods like XGB and RF perform better than the basic methods like DT. Fig.\ref{lbl:unsupervised} shows the AUROC values obtained by unsupervised models, with the RBM, GAN and AE obtain AUROC values above 0.95, while the OC-SVM performs not very well with AUROC = 0.90. Overall, it can be observed that supervised models perform slightly better than unsupervised models, at the expense of additional preprocessing procedures like outliers remove. 

\section{Discussions}
In credit card fraud detection, supervised learning aims to train a binary classification model to distinguish between fraudulent and non-fraudulent instances by feeding labeled data, while unsupervised learning is intended to model data distribution of one class and determine whether a test sample belongs to this class or not. In this section, we will discuss the pros and cons of both supervised and unsupervised learning.

Assuming there are sufficient labeled data, supervised learning models, especially for deep neural networks, are able to achieve very promising classification performance. For example, AlexNet \cite{krizhevsky2012imagenet} significantly reduce error rates for image classification on a large-scale image dataset with more than 1 million labeled images. However, in credit card fraud detection, the training data in two classes are dramatically imbalanced. The fraudulent transactions are much less than the non-fraudulent ones. As a result, the trained classifier will be biased by the majority class whereas it should pay more attention to the minority one. Another issue for supervised learning is that transaction data could only be labeled after several days even a month. This kind of verification latency \cite{krivko2010hybrid} would yield the delay for updating the supervised model. To summarize, the advantage of supervised learning is being capable to achieve very promising results given sufficient training data, while the disadvantage is being dramatically affected by the data imbalance issue and the data labeling processing.

Although unsupervised learning is not so attractive as the supervised one, it is suitable for credit card fraud detection as it does not require balanced label data. For example, the AnoGAN model \cite{schlegl2017unsupervised} is able to learn the normal data distribution and indicate whether an unknown test data is normal or abnormal by using their proposed anomaly scoring scheme. This sort of unsupervised learning model would be more prominent if label data is insufficient and data imbalance is severe. Another advantage for unsupervised learning is that a fraudulent credit card use could be detected promptly because the unsupervised model can be updated in low latency by using online unlabeled data in banks and financial institutes. For example, one of unsupervised learning models, Self-Organizing Map (SOM) \cite{zaslavsky2006credit}, is used to build a framework for unsupervised credit card fraud detection. The proposed automated system is able to continuously modify the model by using new added transactions because the SOM model does not require priori information, e.g., whether a transaction is done by the cardholder or not. In sum, the advantage of unsupervised learning methods are quite obvious for credit card fraud detection, while the disadvantage may be the difficulty of making some unsupervised model (e.g., GAN) converge. 

\section{Conclusions}
In this paper, we conduct a comparison study for credit card fraud detection in a supervised vs unsupervised manner by evaluating $10$ machine learning models on a Kaggle dataset with credit card transactions data. The label availability and the data imbalance restrict the supervised learning performance dramatically, while the unsupervised one does not have these bottlenecks. Moreover, some unsupervised learning methods, e.g., GAN, have recently received more attentions from the community and also achieved very promising results. In futures, we will focus on using GAN models to improve the performance of credit card fraud detection.
\clearpage
\bibliographystyle{aaai}
\bibliography{draft}
\end{document}